\documentclass[runningheads]{llncs}
\usepackage{glossaries}
\usepackage{mathtools}
\usepackage{relsize}
\usepackage{booktabs}
\usepackage{capt-of}
\usepackage{wrapfig}
\usepackage{xspace}

\usepackage{graphicx}
\usepackage{comment}
\usepackage{amsmath,amssymb}
\usepackage{color}
\usepackage{url}
\usepackage[pagebackref=true,breaklinks=true,letterpaper=true,colorlinks,bookmarks=false]{hyperref}
\usepackage[noabbrev,capitalize]{cleveref}
\usepackage[misc]{ifsym}

\newacronym{jsd}{JSD}{Jensen-Shannon Divergence}
\newacronym{dnn}{DNN}{Deep Neural Network}
\newacronym{cnn}{CNN}{Convolutional Neural Network}
\newacronym{dv}{DV}{Donsker \& Varadhan}
\newacronym{iic}{IIC}{Invariant Information Clustering}
\newacronym{miou}{mIoU}{mean Intersection-Over-Union}
\newacronym{mi}{MI}{mutual information}
\newacronym{kl}{KL}{Kullback-Leibler}
\newacronym{arl}{ARL}{Autoregressive Representation Learning}
\newacronym{pa}{PA}{Pixel-Accuracy}
\newacronym{ac}{AC}{Autoregressive Clustering}

\makeatletter
\DeclareRobustCommand\onedot{\futurelet\@let@token\@onedot}
\def\@onedot{\ifx\@let@token.\else.\null\fi\xspace}

\def\eg{\emph{e.g}\onedot} 
\def\ie{\emph{i.e}\onedot}

\def\etal{\emph{et al}\onedot}

\newif\ifreview
\reviewfalse

\ifreview
	\usepackage{lineno}

	\linenumbers
\fi

\begin{document}

%%%%%%%%%%%%%%%%%%%%% Add submission id, track, and title. %%%%%%%%%%%%%%%%%%%%%

% Insert your submission number here
\def\SubNumber{48}

% Choose one track by uncommenting one of the following lines  
\def\GCPRTrack{Regular Track}
% \def\GCPRTrack{Track: Computer vision systems and applications}
% \def\GCPRTrack{Track: Pattern recognition in the life and natural sciences}
% \def\GCPRTrack{Track: Photogrammetry and remote sensing}
% \def\GCPRTrack{Track: Robot vision}
% \def\GCPRTrack{Track: DAGM Young Researcher Forum}

% Replace with your title
\title{InfoSeg: Unsupervised Semantic Image Segmentation with Mutual Information Maximization}
% You can use \thanks for acknowledgment. Do not add any acknowledgment to the draft 
% version that is used for the review process.  
%\title{Title\thanks{XXX}}

\ifreview
	% ANONYMOUS SUBMISSION FOR REVIEW
	% DO NOT MODIFY these for the draft version that is used for the review process.
	\titlerunning{DAGM GCPR 2021 Submission \SubNumber{}. CONFIDENTIAL REVIEW COPY.}
	\authorrunning{DAGM GCPR 2021 Submission \SubNumber{}. CONFIDENTIAL REVIEW COPY.}
	\author{DAGM GCPR 2021 - \GCPRTrack{}}
	\institute{Paper ID \SubNumber}
\else
	\titlerunning{InfoSeg: Unsupervised Semantic Image Segm. with Mutual Information Max.}

	\author{Robert Harb$^{\textrm{(\Letter)}}$  \and
	Patrick Knöbelreiter  }

	\authorrunning{R. Harb and P. Knöbelreiter}
	
	\institute{Institute of Computer Graphics and Vision, Graz University of Technology, Austria \\
    \email{robert.harb@icg.tugraz.at}
    }
\fi

\maketitle              % typeset the header of the contribution

\vspace{-2.0em}
\begin{abstract}
    We propose a novel method for unsupervised semantic image segmentation based on mutual information maximization
    between local and global high-level image features.
    The core idea of our work is to leverage recent progress in self-supervised image representation learning.
    Representation learning methods compute a single high-level
    feature capturing an entire image. In contrast, we
    compute multiple high-level features, each capturing image
    segments of one particular semantic class. To this end,
    we propose a novel two-step learning procedure comprising
    a segmentation and a mutual information maximization
    step. In the first step, we segment images based on local
    and global features. In the second step, we maximize the
    mutual information between local features and high-level
    features of their respective class. For training, we provide
    solely unlabeled images and start from random network initialization.
    For quantitative and qualitative evaluation, we
    use established benchmarks, and COCO-Persons, whereby
    we introduce the latter in this paper as a challenging novel
    benchmark. InfoSeg significantly outperforms the current
    state-of-the-art, e.g., we achieve a relative increase of $26\%$
    in the Pixel Accuracy metric on the COCO-Stuff dataset.

\end{abstract}

\section{Introduction}
\label{sec:intro}
Semantic image segmentation is the task of assigning a
class label to each pixel of an image.
Various applications make use of it, including autonomous driving, augmented reality, or medical imaging.
As a result, a lot of research was dedicated to semantic segmentation in the past.
However, the vast majority of research focused on supervised methods.
A major drawback of supervised methods is that they require large labeled training datasets containing images together with pixel-wise class labels.
These datasets have to be created manually by humans with great effort.
For example, annotating a single image of the Cityscapes \cite{cityscapes} dataset required $90$ minutes of human labor on average.
This dependence of supervised methods on large human-annotated training datasets limits practical applications.
We tackle this problem by introducing a novel approach on semantic image segmentation that does not require any labeled training data.

\begin{figure}[t]
  \includegraphics[width=1.\columnwidth]{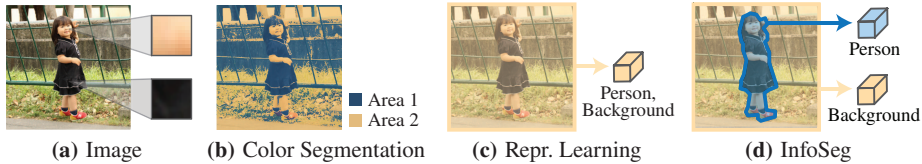}
  \caption{
    \textbf{(a)} Input image. The two magnified image patches have vastly different low-level appearance despite covering the same semantic object: a person.
    \textbf{(b)} Color based segmentation fails to capture any high-level structure of the image.
    \textbf{(c)} Representation learning captures high-level information of the entire image in a single feature.
    \textbf{(d)} InfoSeg captures semantically similar image areas in separate features.
  }
  \label{fig:seg_exp}
\end{figure}

%\begin{figure}
%  \includegraphics[width=1.\columnwidth]{figures/new_teaser.eps}
%  \centering
%  \caption{
%  InfoSeg is trained with unlabeled images and predicts image segmentations.
%  The resulting segmentations capture high-level structures, and directly correspond to semantic classes (person, road, vegetation, \dots).
%  We use solely unlabeled images for training and start from random initialization.
%  \vspace{-1.0em}
%\end{figure}
The major challenge of semantic image segmentation is to identify high-level structures in images.
State-of-the-art methods approach this by learning from labeled data.
While extensive research exists in segmentation without labeled data, it mainly focuses on non-learning based methods using low-level features such as color or edges~\cite{felzenszwalb2004efficient,comaniciu2002mean,ncut,arbelaez2010contour}.
In general, low-level features are insufficient for semantic segmentation.
They are not homogenous across high-level structures.
\Cref{fig:seg_exp}(a-b) illustrate this problem.
An image depicting a person is segmented based on color.
Color changes vastly across image areas, even if they are semantically correlated.
Consequently, the resulting segmentation does not capture any high-level structures.
%For semantic segmentation, we need features that capture high-level image structures.
Contrarily,~\cref{fig:seg_exp}(d) illustrates how InfoSeg maps unlabeled images to segmentations that capture high-level structures.
These segmentations often directly capture the semantic classes of labeled datasets.

The core idea of our method is to leverage image-level representation learning for pixel-level segmentation.
Only recently, self-supervised representation learning methods \cite{deepinfo,simclr,cpc} showed how to extract high-level features from images without any annotated training data.
However, they compute features that capture the \textit{entire} content of images.
Therefore, they are not suitable for segmentation.
To enable segmentation, we instead use multiple high-level features, each capturing semantically similar image areas.
This allows us to assign pixels to classes based on their attribution to each of these features.
\Cref{fig:seg_exp}(c-d) illustrate how our approach differs from image-level representation learning.
We learn high-level features with a \gls{mi} maximization approach, inspired by Local Deep InfoMax \cite{deepinfo}.
However, unlike Local Deep InfoMax, we follow a novel two-step learning procedure enabling segmentation.
At each iteration, we perform a Segmentation and Mutual Information Maximization step.
In the first step, we segment images using the current features.
In the second step, we update the features based on the segmentation from the first step.
This two-step procedure allows us to train InfoSeg using solely unlabeled images and without pre-trained network backbones.

We motivate the exact structure of InfoSeg by first giving a thorough review of current-state-of-the-art methods \cite{Ji_2019_ICCV,Ouali_2020_ECCV}, followed by a discussion of their limitations and how we approach them in InfoSeg.
Our qualitative and quantitative evaluation show that InfoSeg significantly outperforms all compared methods.
For example, we achieve a relative increase of $26\%$ in \gls{pa} on the COCO-Stuff dataset \cite{cocostuff}.
Even though we follow the standard evaluation protocol for quantitative evaluation, we provide a critical discussion of it and uncover problems left undiscussed by recent work \cite{Ji_2019_ICCV,Ouali_2020_ECCV}.
Furthermore, in addition to established datasets, we introduce COCO-Persons as a novel benchmark.
COCO-Persons contains complex scenes requiring high-level interpretation for segmentation.
Our experiments show that InfoSeg handles the challenging scenes of COCO-Persons significantly better than compared methods.
Finally, we perform an ablation study.

\section{Related work}
\paragraph{Self-Supervised Image Representation Learning}
aims to capture high-level content of images without using any labeled training data.
State-of-the-art methods follow a contrastive learning framework \cite{cpc,henaff2019data,deepinfo,bachman19,simclr,dosovitskiy2014discriminative,wu2018unsupervised,he2019momentum}.
In contrastive learning, one computes multiple representations of differently augmented versions of the same input image.
Augmentations can include photometric or geometric image transformations.
During training, one enforces similarity on representations computed from the same image and dissimilarity on representations of different images.
To this end, various objectives exist, such as the normalized cross entropy \cite{cpc} or \gls{mi} \cite{deepinfo}.

%Infoseg is based on Local Deep InfoMax \cite{deepinfo}.
%Local Deep InfoMax computes a single high-level representation capturing an entire image.
%In contrast, we compute multiple representations, each capturing one image segment.
%The authors of AMDIM \cite{bachman19} also proposed to compute multiple representations from images in their mixture-based representations.
%However, they did not tackle semantic image segmentation and provided only coarse low-resolution attention maps.
%Contrarily, our approach outputs segmentations at the resolution of input images.

%\vspace{-.6em}
%\paragraph{Supervised Semantic Image Segmentation.}
%Various approaches exist that require annotated training data. Supervised methods \cite{long2015fully,chang2018pyramid,chen2017rethinking} require pixel-wise annotations.
%Weakly-supervised methods require coarse annotations such as bounding boxes (location and scale of segments) \cite{song2019box,dai2015boxsup,khoreva2017simple,redondo2019learning,papandreou2015weakly},
%scribbles (partially pixel-wise annotation of segments) \cite{lin2016scribblesup,tang2018normalized,tang2018regularized,marin2019beyond}
%or image-level annotations \cite{lee2019ficklenet,kim2017two,wei2018revisiting,ahn2018learning}.
%Semi-supervised methods require a small dataset with pixel-wise annotations and a large unlabeled dataset \cite{Hung2018AdversarialLF,souly2017semi,kalluri2019universal}.
%However, only a small amount of research deals with semantic image segmentation without any annotated training data.
%We briefly review unsupervised methods in the following.

\paragraph{Unsupervised Semantic Image Segmentation.}
\gls{iic} \cite{Ji_2019_ICCV} is a clustering approach also applicable for semantic segmentation.
Briefly, \gls{iic} uses a \gls{mi} objective that enforces the same prediction for differently augmented image patches.
The authors of \gls{iic} proposed to use photometric or geometric image transformations to compute augmentations.
For example, one can create augmentations by random color jittering, rotation, or scaling.
Ouali \etal \cite{Ouali_2020_ECCV} did a follow-up work on \gls{iic}.
In addition to standard image transformations, they proposed to process image patches through various masked convolutions.
We further discuss these two methods and its differences to InfoSeg in~\cref{sec:moti}.
Concurrent to our work, Mirsadeghi \etal proposed InMARS \cite{mirsadeghi2021unsupervised}.
InMARS is also related to IIC. However, instead of operating on each pixel individually, InMARS utilizes a superpixel representation.
Furthermore, a novel adversarial training scheme is introduced.

%To provide additional baseline models, the authors of IIC applied K-Means clustering on patch-wise
%image features learned by representation learning methods.
%In particular, DeepCluster \cite{deepcluster} and methods from Doersch et al. \cite{doersch2015unsupervised} and Isola et al. \cite{isola2015learning}.
%However, since these methods are not designed for semantic image segmentation, the resulting segmentations are not competitive with dedicated methods.

Another recently introduced method that states to perform unsupervised semantic segmentation is SegSort \cite{Hwang_2019_ICCV}.
%SegSort uses an Expectation-Maximization framework to map each pixel of an input image into a feature space.
%It then assigns pixels that are close in this space to the same segment.
However, we note that SegSort still uses supervised learning at multiple stages.
First, they initialize parts of their network architecture with pre-trained
weights obtained by supervised training of a classifier on the ImageNet \cite{imagenet} dataset.
Second, they use pseudo ground truth masks generated by a
HED contour detector \cite{holedge}, which is trained supervised using the BSDS500 \cite{arbelaez2010contour} dataset.
Therefore, we do not consider SegSort as an unsupervised method.

\section{Motivation}
\label{sec:preliminaries}
In this section, we first review how recent work \cite{Ji_2019_ICCV,Ouali_2020_ECCV} uses \gls{mi} for unsupervised semantic image segmentation.
Then, we discuss limitations of these methods, and how we tackle them in InfoSeg.

\subsection{Unsupervised Semantic Image Segmentation}
State-of-the-art methods \cite{Ji_2019_ICCV,Ouali_2020_ECCV} adapt the \gls{mi} based image clustering approach of \gls{iic} \cite{Ji_2019_ICCV} for segmentation.
In the following, we introduce IICs' approach on image clustering and then the proposed modifications for segmentation.

For clustering, one creates two versions $x$ and $x'$ of the same image.
These versions show the same semantic content, but alter low-level appearance by using random photometric or geometric transformations.
Consequently, semantic class predictions $y$ and $y'$ of the two images $x$ and $x'$ should be the same.
To achieve this, one maximizes the \gls{mi} between $y$ and $y'$
\begin{equation}
    \max_{\psi} \quad
    I(\Phi_{\psi}(x); \Phi_{\psi}(x')) = I(y;y'),
    \label{eq:ic}
\end{equation}
where $\Phi$ is a CNN parametrized by $\psi$.
Considering we can express the \gls{mi} between $y$ and $y'$ as
\begin{equation}
    I(y;y') = H(y) - H(y|y'),
    \label{eq:mient}
\end{equation}
\cref{eq:ic} maximizes the entropy $H(y)$ while minimizing the conditional entropy $H(y|y')$.
Minimizing $H(y|y')$ pushes predictions of the two images $x$ and $x'$ together.
Therefore, the network has to compute predictions invariant to the different low-level transformations.
This should encourage class predictions to depend on high-level image content instead.
While sole minimization of $H(y|y')$ can trivially be done by assigning the same class to all images.
Additional maximization of $H(y)$ has a regularization effect against such degenerate solutions.
Since maximizing $H(y)$ encourages predictions that put equal probability mass on all classes.
Consequently, predictions for all images can not collapse to a single class.

For segmentation, Ji \etal \cite{Ji_2019_ICCV} proposed to use the previously introduced clustering approach on image patches rather than entire images.
Two image versions are pushed through a network that computes dense pixel-wise class predictions.
The objective given in~\cref{eq:ic} is now applied on the pixel-wise class predictions.
Therefore, each prediction depends on an image patch rather than an entire image.
Patches are defined by the receptive field for each output pixel of the network.
Additionally, one enforces local spatial invariance by maximizing \gls{mi} of predictions from adjacent image patches.
This approach on unsupervised semantic segmentation was initially proposed by \gls{iic} \cite{Ji_2019_ICCV}.
Furthermore, Ouali \etal \cite{Ouali_2020_ECCV} proposed an extension by generating views using different masked convolutions \cite{van2016conditional}.
In the following, we discuss three major limitations of these two works, and how we tackle them in InfoSeg.
%For example $g$ can consist of . %like random jitter of image contrast, saturation or brightness and like rotation, scaling, skewing or flipping.
%Given that $y,y' \in [0,1]^K$ are discrete $K$ dimensional random variables modeling the pixel-wise assignment to one out of $K$ classes.

%\begin{figure}[t]
%    \begin{center}
%    \includegraphics[width=1.15\columnwidth]{figures/iic_comp.eps}
%    \caption{IIC}
%    \label{fig:iic}
%    \end{center}
%\end{figure}

\subsection{Limitations of current methods}
\label{sec:moti}
The first limitation of discussed methods is that they do not incorporate global image context.
Global context is essential to capture high-level structures, since they often cover large image areas having diverse local appearance.
Therefore observing only small image patches is often not sufficient to identify them.
Ideally, each pixel-wise prediction should depend on the entire image.
Nevertheless, the discussed approaches make pixel-wise predictions based on image patches.
The receptive field of the network $\Phi$ determines the size of these patches.
In general, one could enlarge the receptive field by changing the network architecture.
However, adapting \gls{iic} from clustering to segmentation is based on restricting each prediction's receptive field from entire images to patches.
By making each pixel-wise prediction dependent on the entire image again, one would fall back to clustering.
In InfoSeg we capture global context in global high-level features that cover the entire image.
We make pixel-wise predictions based on the \gls{mi} between these global features and local patch-wise features.
This allows each pixel-wise prediction to depend on the entire image.

A second limitation of discussed methods is that they fail to leverage recent advances in image representation learning \cite{deepinfo,simclr,cpc,bachman19}. These methods are effective at capturing high-level image content, but only at the image-level.
Adapting them for pixel-level segmentation is not trivial.
Ouali \etal \cite{Ouali_2020_ECCV} attempted this with their \gls{arl} loss, but failed to increase segmentation performance.
Despite high-level information is constant across large image areas,
\gls{arl} computes for \textit{each pixel} a separate high-level feature.
Contrarily, in InfoSeg, we share high-level features over the \textit{entire image}.
To still allow pixel-wise segmentation, we compute multiple high-level features.
Each high-level feature encodes only image areas depicting one class.
We then assign pixels to classes based on their attribution to each of these features.

Finally, discussed methods jointly learn features and segmentations.
They use intermediate feature representations to assign pixels to class labels.
At the beginning of training, features depend on random initialization and contain no high-level information.
This can lead to classes that latch onto low-level features instead of capturing high-level information.
This issue was first discussed for image classification by SCAN \cite{wvangansbeke2020learning}.
Instead, we decouple feature learning and segmentation.
Therefore, we perform two steps at each iteration.
First, we compute features that are explicitly trained to encode high-level information.
Then, we use them for segmentation.

\begin{figure*}[t!]
    \centering
    \includegraphics[width=1.0\columnwidth]{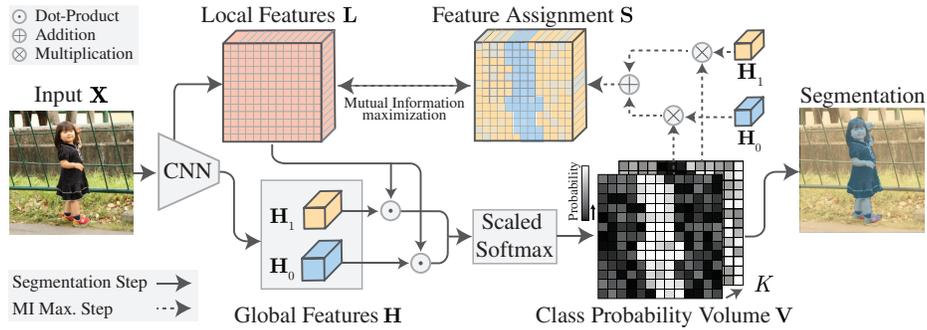}
    \caption{Overview of InfoSeg for $K=2$ classes.
    At each training iteration, we alternate the following two steps.
    \textbf{Segmentation Step} (solid lines): An input image $\mathbf{X}$ is passed through a CNN to compute local patch-wise features $\mathbf{L}$ and for each class $k$ a global image-level feature $\mathbf{H}_k$.
    We then score local $\mathbf{L}$ with global $\mathbf{H}$ features using a dot-product. The result is passed through a scaled softmax function to compute the class probability volume $\mathbf{V}$.
    Finally, we obtain a segmentation by assigning each pixel to the class with the largest probability.
    \textbf{Mutual Information Maximization Step} (dashed lines):
    The global feature assignment $\mathbf{S}$ is computed as a sum of global features, weighted by their respective class probabilities at each spatial position.
    Finally, we maximize Mutual Information between local features $\mathbf{L}$ and their respective feature assignment $\mathbf{S}$.
    }
    \label{fig:overview}
\end{figure*}

\section{InfoSeg}
\label{sec:infoseg}

In InfoSeg, we tackle unsupervised semantic image segmentation.
We take a set $\{\mathbf{X}^{(n)} \in \mathcal{X}\}_{n=1}^N$ of $N$ unlabeled images and assign a label $\mathcal{Z} = \{z_1, \dots, z_{K}\}$ to every pixel of each image.
Importantly, for one particular image, we do not specify which nor how many labels should be assigned.
We only provide the total number of labels $K$ in all images.
After training, we follow the standard evaluation protocol and map the learned labels of InfoSeg directly to the semantic classes of an annotated dataset.
%A single global feature capturing a high-level representation of an entire image can be learned
%unsupervised by maximizing its average mutual information with local patch-wise features~\cite{deepinfo}.

InfoSeg is designed to tackle the three limitations of state-of-the-art methods discussed in~\cref{sec:moti}.
\Cref{fig:overview} shows an overview of InfoSeg.
%We use global high-level features capturing image segments to perform semantic image segmentation.
%To learn these features we follow recent methods \cite{deepinfo, bachman19}.
%However, while these methods operate at the image-level, we tackle pixel-level segmentation.
In the following, we first discuss how we leverage recent progress in representation learning for semantic segmentation in~\cref{sec:murep}.
Then we provide further details of our method in~\cref{sec:seg} and~\cref{sec:mumax}.

\subsection{Representation Learning for Segmentation}
\label{sec:murep}

%One approach on this is to maximize Mutual Information
%between images and their respective representations $I(X;H)$.
%However this objective can simply be maximized if pixel-level noise
%of images is encoded in $H$.
%Hence, these representations have shown to be less suitable for tasks
%requiring high-level information e.g. classification \cite{deepinfo}.
We first review how Local Deep InfoMax \cite{deepinfo} captures high-level information of entire images,
and then how InfoSeg adapts this approach to target image segmentation.

Local Deep InfoMax \cite{deepinfo} learns global high-level features of images by maximizing their average \gls{mi} with local features.
Local features cover image patches, and the global feature covers the entire image.
If the global feature has limited capacity, the network cannot simply copy all local features' content into the global feature to maximize \gls{mi}.
Instead, the network has to encode a compact representation that shares information with as many image patches as possible.
Hjelm \etal \cite{deepinfo} showed that the resulting global features encode high-level image information.
They motivated this by the idea that high-level information is often constant over an entire image, while low-level information such as pixel-level noise varies.
Consequently, the global feature is encouraged to encode the former while disregarding the latter.

To enable pixel-wise segmentation, we compute for each image multiple global features instead of a single one.
Each global feature only encodes image areas that depict a particular class.
This allows us to segment images by assigning pixels to classes based on their attribution to each global feature.
During training, we maximize for each global feature \gls{mi} only with local features covering its respective class.
Therefore, we learn high-level features in a similar way as Local Deep InfoMax \cite{deepinfo}, but target segments instead of entire images.
This requires us to learn high-level features together with segmentations.
To this end, we alternate two steps at each iteration.
In the \textit{Segmentation Step}, we assign local to global features based on their content, \ie, we segment images.
In the \textit{Mutual Information Maximization step}, we maximize the \gls{mi} between all global features and assigned local features, \ie, we learn the features.
We describe both steps in the following.

\begin{figure}[t]
	\begin{center}
	\includegraphics[width=.85\columnwidth]{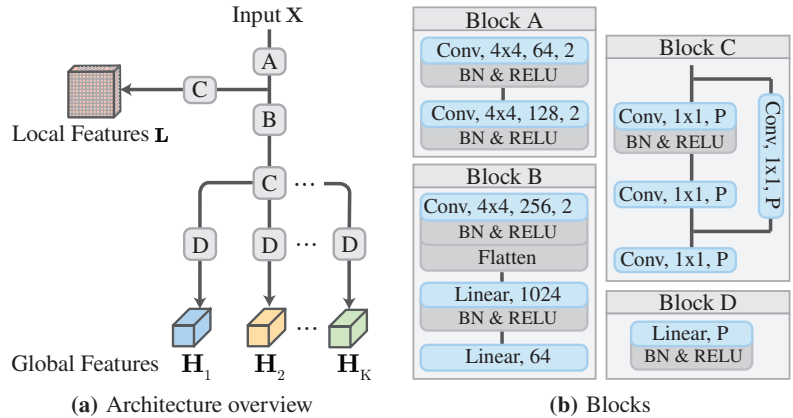}
	\caption{Feature computation for $K$ classes. \textbf{(a)} Overview of network architecture.
		\textbf{(b)} Used blocks. \textbf{Legend}: \textit{Conv, $W\!\times\!W,C,d$}: Convolution with filter size $W\!\!\times\!W$, $C$ channels and stride $d$.
      Blocks that are used multiple times, each have their own set of parameters.}
	\label{fig:blocks}
	\end{center}
\end{figure}

\subsection{Segmentation Step}
\label{sec:seg}
Given an input image $\mathbf{X} \in \mathcal{X} = \mathbb{R}^{M \times N \times C}$, we compute $P$-dimensional global $\mathbf{H} \in \mathbb{R}^{K \times P}$ and patch-wise local $\mathbf{L} \in \mathbb{R}^{U \times V \times P}$ features.
The $k$-th global feature $\mathbf{H}_k \in \mathbb{R}^{P}$ encodes a high-level representation for the $k$-th class and covers the entire image.
The local feature $\mathbf{L}_{i,j} \in \mathbb{R}^{P}$ at the spatial position $(i,j)$ encodes an image patch.
Furthermore, the spatial resolution of $\mathbf{L}$ is downsampled by a rate of $d$ from the input resolution, \ie $U=M/d$ and $V=N/d$.

\Cref{fig:blocks} shows the architecture of our feature computation network.
First, the input image is processed by Block A, resulting in a grid of patch-wise image features.
To compute the local features $\mathbf{L}$, we further process these patch-wise features by Block C.
Adding this additional residual block of pointwise convolutions led to better performance, than using Block A's output directly for the local features.
To compute the global features $\mathbf{H}_k$, we first process the output of Block A to image-level features using Block B.
Then, similarly, as for the local features, we add a residual block of pointwise convolutions using Block C.
Finally, each global feature is computed using a separate linear layer using Block D.

To compute an image segmentation, we use the dot-product of a local and global feature pair $\langle \mathbf{L}_{i,j}, \mathbf{H}_k \rangle$ as a class score.
A high score indicates that the $k$-th class is shown at the position $(i,j)$.
We elaborate in~\cref{sec:mumax} how \gls{mi} maximization increases the dot-product of a local feature and the global feature of its corresponding class.
After computing the class scores, we apply a pixel-wise scaled softmax to compute a class-probability volume $\mathbf{V} \in \mathbb{R}^{U \times V \times K}$ with elements
\begin{equation}
  V_{i,j,k} = \frac{\exp \left(\tau \cdot \langle \mathbf{L}_{i,j}, \mathbf{H}_k \rangle \right)}{\sum_{\hat{k}}
  \exp \left(\tau \cdot \langle \mathbf{L}_{i,j}, \mathbf{H}_{\hat{k}} \rangle \right)},
  \label{eq:prob-vol}
\end{equation}
where $\tau$ is a hyper-parameter that controls the smoothness of the resulting distribution.
Using the probability volume, we compute the low-res segmentation $\mathbf{K}$ for every pixel $(i,j)$ with
\begin{equation}
  k_{i,j} = \operatorname*{arg\,max}_{k \in \mathcal{Z}} V_{i,j,k},
  \label{eq:class}
\end{equation}
by taking the class with the largest probability.
We can then compute a full-res segmentation $\mathbf{Z}$ by upsampling the low-res segmentation $\mathbf{K}$ to the input image resolution. % by

\subsection{Mutual Information Maximization Step}
\label{sec:mumax}
We first need to assign each local feature to its corresponding class's global feature.
We could do this using the segmentation $\mathbf{K}$.
However, this disregards class probabilities, instead of utilizing their exact values, \eg to account for uncertainty.
Especially at the beginning of training, segmentations are uncertain and depend on random network initialization.
Reinforcing possibly incorrect predictions can lead to degenerate solutions.
To alleviate this problem, we do not make hard class assignments using $\mathbf{K}$, but soft assignments using class-probabilities $\mathbf{V}$.
Instead of assigning a single global feature, we weight each global feature by its respective class probability.
To this end, we define the function $S^{(i,j)}_{\boldsymbol{\theta}}$ that computes a soft global feature assignment for the local feature $\mathbf{L}_{(i,j)}$ as follows
\begin{equation}
  S^{(i,j)}_{\boldsymbol{\theta}}(\mathbf{X}) = \sum_k V_{i,j,k} \cdot
  H^{(k)}_{\boldsymbol{\theta}}(\mathbf{X}),
\end{equation}
where the function $H_{\boldsymbol{\theta}}^{(k)}(\mathbf{X})$ computes the $k$-th global feature $\mathbf{H}_{k}$ for an image $\mathbf{X}$, and $\boldsymbol{\theta}$ denotes the learnable parameters of our network.

During training, we maximize the \gls{mi} between the output of $S_{\boldsymbol{\theta}}^{(i,j)}(\mathbf{X})$ and the corresponding local feature $\mathbf{L}_{(i,j)}$ for all spatial positions $(i,j)$.
Hence our objective is given as
\begin{equation}
  \operatorname*{max}_{\boldsymbol{\theta}} \, \mathbb{E}_{\mathbf{X}} \left[
  \frac{1}{UV} \mathlarger{\mathlarger{\sum}}_{i,j}
  I \left( L_{\boldsymbol{\theta}}^{(i,j)}(\mathbf{X}) ;
  S^{(i,j)}_{\boldsymbol{\theta}}(\mathbf{X}) \right) \right],
  \label{eq:trainingobjsoft}
\end{equation}
where $\mathbb{E}_\mathbf{X}$ denotes the expectation over all training images $\mathbf{X}$ and the function $L_{\boldsymbol{\theta}}^{(i,j)}(\mathbf{X})$ computes the local features $\mathbf{L}_{i,j}$ given an input image $\mathbf{X}$.

To evaluate our objective~\cref{eq:trainingobjsoft}, consider that local and global features are high-dimensional continuous random variables.
\Gls{mi} computation of such variables is challenging.
Contrarily to discrete variables as in the objective of \gls{iic}~\cref{eq:ic}, where exact computation is possible.
%For continuous variables, one can estimate \gls{mi} by maximizing lower bounds \cite{mine}.
%Although several bounds exist \cite{varbounds}, we focus on a bound based on the \gls{jsd} \cite{nowozin2016f}.
%Mainly because Hjelm \etal \cite{deepinfo} showed favorable properties of the \gls{jsd} bound compared to others.
%This includes more stable training and better performance with smaller batch sizes.
For continuous variables, Belghazi \etal \cite{mine} proposed \gls{mi} estimation by maximizing lower bounds parametrized by neural networks.
They used a bound based on the \gls{dv} representation of the \gls{kl} divergence.
While several other bounds exist \cite{varbounds}, we use a bound based on the \gls{jsd}.
Mainly because Hjelm \etal \cite{deepinfo} showed favorable properties of the \gls{jsd} bound compared to others in their representation learning setting.
This includes increased training stability and better performance with smaller batch sizes.
Nevertheless, we also perform experiments using the \gls{dv} bound in our ablation studies.
A \gls{jsd} based \gls{mi} estimator $\widehat{I}_{JSD}(X;Y)$ for two random variables $X$ and $Y$ can be defined as follows \cite{nowozin2016f}
\begin{equation}
  {I}(X;Y)
  \geq
  \widehat{I}_{JSD}(X;Y) :=
  \mathbb{E}_{p(x, y)}[-\operatorname{sp}(-T(x, y))]  -\mathbb{E}_{p(x) p(y)}[\operatorname{sp}(T(x, y))],
  \label{eq:muinf}
\end{equation}
where $\mathrm{sp}(x)=\log \left(1+e^{x}\right)$ and $T$ is a discriminator mapping sample pairs from $X$ and $Y$ to a real valued score.
The first and second expectations are taken over samples from the joint $p(x,y)$ and marginal $p(x)p(y)$ distributions.
Consequently, to tighten the bound, the discriminator $T$ needs to discriminate samples from the joint and marginal distributions by assigning high or low scores, respectively.

To use the \gls{jsd} estimator~\cref{eq:muinf} in our objective~\cref{eq:trainingobjsoft}, we have to define the discriminator $T$ and a sampling strategy.
Following recent work \cite{deepinfo,bachman19}, we create joint and marginal samples by combining feature pairs computed from the same image $\mathbf{X}$ and two randomly paired images $\mathbf{X}$ and $\mathbf{X'}$, respectively.
The discriminator $T$ can be implemented using any arbitrary function that maps feature pairs to a discrimination score, \eg, a neural network.
For efficiency, we use the dot-product to compute discrimination scores \ie, $T(x,y):=\langle x, y \rangle$.
This requires only a single expensive forward pass through our network to compute the features, while we can then score any arbitrary combination with a cheap dot-product.
Omitting the spatial indices $(i,j)$ to avoid notational clutter, this leads to the \gls{mi} estimator
\begin{equation}
  \begin{gathered}
    \widehat{I}_{JSD} \left( L_{\boldsymbol{\theta}}(\mathbf{X}) ; S_{\boldsymbol{\theta}}(\mathbf{X}) \right) :=
    \mathbb{E}_{\mathbb{P}}[-\operatorname{sp}(-
    \langle L_{\boldsymbol{\theta}}(x), S_{\boldsymbol{\theta}}(x)\rangle )]-
    \mathbb{E}_{\mathbb{P}\times \mathbb{\tilde{P}}}[\operatorname{sp}(
    \langle L_{\boldsymbol{\theta}}(x) ; S_{\boldsymbol{\theta}}(x') \rangle
    ],
  \end{gathered}
  \label{eq:estim}
\end{equation}
where $\mathbb{P}$ is the empirical distribution of our dataset, $x$ is an image sampled from $\mathbb{P}$ and $x'$ is an image sampled from $\mathbb{\tilde{P}} = \mathbb{P}$.
We can now simply insert the estimator of~\cref{eq:estim} into our objective~\cref{eq:trainingobjsoft}.
Maximizing the resulting objective increases the dot-product of local features with the global feature of their assigned class.
Consequently, we use the dot-product as a class score, as described in~\cref{sec:seg}.

\section{Experiments}
We first introduce our experimental setup and discuss challenges at the quantitative evaluation of unsupervised segmentation.
Then we perform an evaluation using established benchmarks \cite{potsdam,cocostuff}, and COCO-Persons, a novel dataset introduced in this work.
On all datasets, InfoSeg significantly outperforms compared methods.
Finally, we perform ablation studies.

\subsection{Setup}

We start training from random network initialization and provide solely unlabeled images.
We set $P=1024$, $\tau=0.8$ and use the ADAM optimizer \cite{kingma2014adam} with a learning rate of $10^{-4}$ and a batch size of $64$.
Furthermore,
the network architecture we use results in a downsampling rate of $d=4$, and we set the number of classes $K$ to be equal to the number of classes in each dataset.

Note that InfoSeg requires a network with a different structure as \glsfirst{iic} and \glsfirst{ac}.
For InfoSeg, the final outputs are $1\!\!\times\!\!1$ sized global image features.
Contrarily, in \gls{iic} and AC, the final outputs are pixel-wise class predictions downscaled from the input image resolution.
This impedes a comparison with these methods using the exact same architecture.
Nevertheless, we provide an experiment in our ablation study where we apply the objective of \gls{iic} on the output of our Segmentation Step.

%\addtolength{\tabcolsep}{-5pt}
%\begin{tabular}{lc}
%\cline{2-2}
%Ceiling    & 0  \\ \cline{2-2}
%Floor      & 0  \\ \cline{2-2}
%Food      & 0  \\ \cline{2-2}
%Furniture  & 0  \\ \cline{2-2}
%Rawmat.    & 0  \\ \cline{2-2}
%Textile    & 0  \\ \cline{2-2}
%Wall       & 16 \\ \cline{2-2}
%Window     & 0  \\ \cline{2-2}
%Building   & 0  \\ \cline{2-2}
%Ground     & 36 \\ \cline{2-2}
%Plant      & 66 \\ \cline{2-2}
%Sky        & 81 \\ \cline{2-2}
%Solid      & 0  \\ \cline{2-2}
%Structural & 8  \\ \cline{2-2}
%Water      & 49 \\ \cline{2-2}
%\end{tabular}

\subsection{Quantitative Evaluation}
\label{sec:quanti}
Meaningful quantitative evaluation of unsupervised semantic segmentation is challenging.
Recent work used the \gls{pa} for quantitative evaluation.
The \gls{pa} is defined as the percentage of pixels assigned to the same class as in a given annotation.
However, in unsupervised semantic segmentation, one does not specify which classes should be used for segmentation.
Instead, many different segmentations can be considered as equally valid.
Nevertheless, quantitative evaluation metrics, such as the \gls{pa} or \gls{miou}, evaluate all pixel-wise predictions as incorrect that do not exactly match the given annotations.
%\Cref{fig:quant}(c) illustrates this problem.
%Nevertheless large portions are evaluated as incorrect by the \gls{pa} since they do not match the annotation.
%Our prediction assigns the person and the motorbike to the same segment.
%Contrarily, they are both assigned to two different segments in the annotations of the COCO-Persons dataset.
%Even though our prediction differs from the annotation and is evaluated as incorrect by the \gls{pa}, it is still a reasonable segmentation.
While this has been left undiscussed by previous work \cite{Ji_2019_ICCV,Ouali_2020_ECCV},
we emphasize this has to be considered when interpreting quantitative metrics of unsupervised methods.

We can further illustrate problems at quantitative evaluation using the COCO-Stuff \cite{cocostuff} dataset as an example.
The dataset contains the class \textit{rawmaterial} that labels image areas depicting metal, plastic, paper, or cardboard.
We argue that this is a very specific class and aggregating these four materials in one class is an arbitrary design choice of the dataset.
It is unfeasible to expect an unsupervised method to come up with this specific solution.
Nevertheless, recent methods \cite{Ji_2019_ICCV,Ouali_2020_ECCV} reported significant increases over baseline models on the \gls{pa}.
We attribute this to the dataset's vast class imbalance.
Besides very specific classes such as \textit{rawmaterial}, the dataset also contains more generic classes such as \textit{water}, or \textit{plant}.
These classes are overrepresented and make up more than $50\%$ of all pixels.
Therefore, an algorithm can achieve high \gls{pa} by focusing mainly on these few overrepresented classes.
To illustrate this effect, we provide a confusion matrix of our predictions in the supplementary material.

Despite the discussed problems, we follow prior work and use the \gls{pa} to evaluate all of our results quantitatively.
Following the standard evaluation protocol \cite{Ji_2019_ICCV,Ouali_2020_ECCV}, we map each of the predicted classes in $\mathcal{Z}$ to one of the annotated classes in $\mathcal{Z'}$ before computing the \gls{pa}.
This is necessary because class ordering is unknown without providing labeled data during training.
We find the one-to-one mapping between $\mathcal{Z}$ and $\mathcal{Z}'$ by solving the linear assignment problem using the Hungarian method \cite{kuhn1955hungarian}.
We compute this mapping once after training and use the same mapping for all images in the dataset.

%\begin{table}
%    \footnotesize
%    \centering
%\end{table}

\interfootnotelinepenalty=10000
\subsection{Data}
\label{sec:data}
Recent work \cite{Ji_2019_ICCV,Ouali_2020_ECCV} established the COCO-Stuff \cite{cocostuff} and Potsdam \cite{potsdam} datasets as benchmarks.
COCO-Stuff contains 15 classes and Potsdam 6 classes.
Additionally, for both datasets, a reduced 3-class variation exists.
We use the same pre-processing as in the compared methods, resulting in $128\!\times\!128$ sized RGB images for COCO-Stuff, and $200\!\!\times\!\!200$ sized RGBIR images for Potsdam.

%The overrepresented classes in COCO-Stuff mainly have a homogeneous low-level appearance.
%For example, color and texture are often sufficient to segment areas labeled as \textit{water}.
%While this is still challenging in the unsupervised setting, we show that InfoSeg can go one step further by segmenting more complex scenes.
%To this end, we introduce the COCO-Persons dataset in the following.

%COCO-Persons is a novel subset of the COCO dataset \cite{cocodataset}
%Each image depicts one or multiple persons and is annotated with a person and a non-person class.
%Unlike some classes in COCO-Stuff, these two classes are not overly specific, and their pixel-frequency over the entire dataset is approximately equal.
%Furthermore, images in the dataset have a heterogeneous low-level appearance.
%Face, hair, and clothing of persons vary vastly in color, texture, and shape, and the non-person areas cover a variety of complex indoor and outdoor scenes.
%Therefore, computing valid semantic segmentations is more challenging than for the overrepresented classes in COCO-Stuff.
%The dataset contains $15\,399$ images resized to $128\!\!\times\!\!128$ pixels.

While unsupervised segmentation of COCO-Stuff and Potsdam is challenging, most classes in these datasets still have a homogeneous low-level appearance.
For example, low-level features such as color and texture are often sufficient to segment areas labeled as \textit{water} in COCO-Stuff or \textit{road} in Potsdam.
To show that InfoSeg can go one step further, we evaluate on an additional dataset where 
segmentation is more reliant on high-level image features.
To this end, we introduce the COCO-Persons dataset, which we will provide publicly.
Each image depicts one or multiple persons and is annotated with a person and a non-person class.
Face, hair, and clothing of persons vary vastly in color, texture, and shape, and the non-person areas cover a variety
of complex indoor and outdoor scenes.
The dataset is a subset of the COCO \cite{cocodataset} dataset and contains $15\,399$ images having $128\!\!\times\!\!128$ pixels.

\begin{figure*}[t]
    \centering
    \includegraphics[width=1.0\columnwidth]{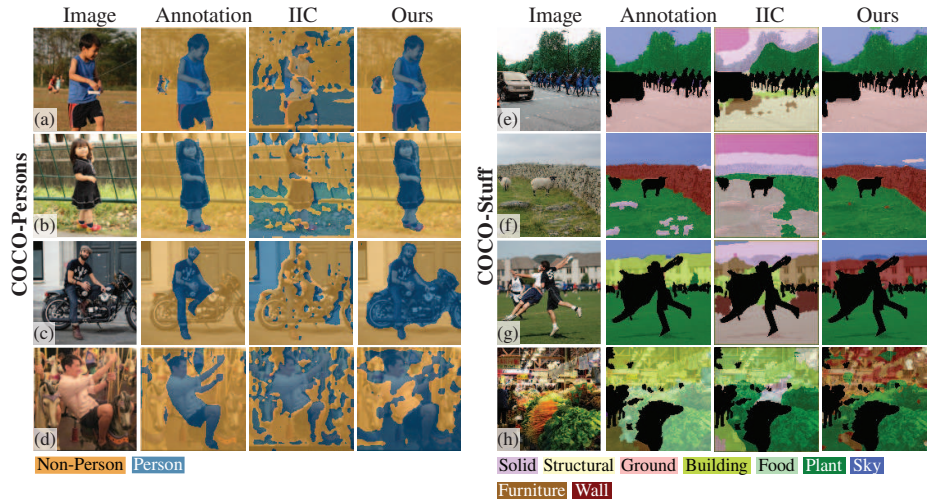}
    \caption{Qualitative comparison. Non-stuff areas in COCO-Stuff are masked in black.}
    \label{fig:quant}
\end{figure*}

\subsection{Results}
\begin{table*}[t!]
    \small
    \centering
    \begin{tabular}{@{}lcccccc@{}}
        Method & COCO-Persons & COCO-Stuff & COCO-Stuff-3 & Potsdam & Potsdam-3 \\
        \midrule
        Random CNN &  $52.3$ & $19.4$  &   $37.3$ & $28.3$  &  $38.2$               \\
        K-Means &  $54.3$ & $14.1$     &  $52.2$  & $35.3$   &  $45.7$           \\
        Doersch$^*$ \cite{doersch2015unsupervised} &  $55.6$ &  $23.1$   &  $47.5$   & $37.2$  &  $49.6$             \\
        Isola$^*$ \cite{isola2015learning} & $57.5$  &  $24.3$         &  $54.0$   & $44.9$  &   $63.9$      \\
        \gls{iic} \cite{Ji_2019_ICCV} & $57.1$          &    $27.7$        & $72.3$     &  $45.4$ &    $65.1$    \\
        AC \cite{Ouali_2020_ECCV} & - & $30.8$& $72.9$& $49.3$  & $66.5$ \\
        InMARS \cite{mirsadeghi2021unsupervised} & - & $31.0$ & $73.1$ & $47.3$ & $70.1$ \\
        \midrule
        InfoSeg (ours)    & $\mathbf{69.6}$ & $\mathbf{38.8}$  &  $\mathbf{73.8}$& $\mathbf{57.3}$  & $\mathbf{71.6}$ \\ \bottomrule
    \end{tabular}
    \vspace{.25em}
    \caption{Pixel-Accuracy of InfoSeg and compared methods.
        $^*$Clustering of features from methods that are not specifically designed for image segmentation.}
    \label{tb:quant}
\end{table*}

\label{sec:res}
We provide quantitative and qualitative results in~\cref{tb:quant} and~\cref{fig:quant}, respectively.
To compute results for COCO-Persons, we used publicly available implementations, if available.
In our experiments, InfoSeg significantly outperformed all compared methods \cite{doersch2015unsupervised,isola2015learning,Ji_2019_ICCV,Ouali_2020_ECCV,mirsadeghi2021unsupervised}.
We discuss qualitative results in the following.
%We provide a detailed discussion of our results in the following.

%\begin{figure}[t]
%  \centering
%  \includegraphics[width=1.0\columnwidth]{figures/conf_mat.eps}
%  \caption{Confusion Matrix for COCO-Stuff. Values are normalized over rows and given in percent.}
%  \label{fig:conf}
%\end{figure}

\paragraph{COCO-Persons.}
%The segmentation of all images requires a high-level interpretation of the depicted scenes, and IIC fails to create any reasonable segmentations.
In~\cref{fig:quant}(a-c), we show successful segmentation of images with vastly inhomogenous low-level appearance.
InfoSeg even captures the two small persons in the background of~\cref{fig:quant}(a).
In~\cref{fig:quant}(c), the motorbike is assigned to the same class as the person.
The dataset contains several images where persons are shown together with motorbikes.
Therefore, without supervision, it is challenging to disentangle these two semantic concepts.
In~\cref{fig:quant}(d), we show a challenging example yielding a failure case.
%In all shown examples IIC fails to create reasonable segmentations, we conclude that our representation learning

\paragraph{COCO-Stuff.}
\Cref{fig:quant}(e-f) show examples where our predictions are close to the annotations.
\Cref{fig:quant}(g-h) provide reasonable segmentations, even though large portions differ from the annotations.
These examples demonstrate challenges at the evaluation of COCO-Stuff due to overly specific classes.
Matching the annotations requires precise distinction of similar high-level concepts, which is difficult without supervision.
The example in~\cref{fig:quant}(g) shows multiple houses
that are assigned to the same class as the stone wall in~\cref{fig:quant}(f).
However, the ground truth of COCO-Stuff assigns the stone wall to a \textit{wall} class and the houses to a \textit{building} class.
\Cref{fig:quant}(h) shows a market scene containing vegetables labeled as \textit{food} but predicted as \textit{plants}.
Arguably, vegetables are food and plants.

\subsection{Ablation Studies}
\setlength{\intextsep}{10pt}
\begin{wraptable}{r}{45mm}
    \small
    \begin{tabular}{@{}lll@{}}
        Measure & MI Max. Step & \gls{pa} \\ \midrule
        IIC-MI  & -                            & 60.2               \\
        DV      & Hard Assignment              & 53.9               \\
        DV      & Soft Assignment              & 55.1              \\
        JSD     & Hard Assignment              & 67.3               \\
        JSD     & Soft Assignment              & \textbf{73.8}               \\
        \bottomrule
    \end{tabular}
    \vspace{-1.5em}
    \caption{Ablation studies on COCO-Stuff-3.}
    \label{tb:ablation}
\end{wraptable}

To examine the influence of individual components, we perform the following ablation studies.
First, we evaluate the effectiveness of soft assignments by replacing them with hard assignments.
Therefore, we change our objective~\cref{eq:trainingobjsoft} to maximize the \gls{mi} at each spatial position between the local feature and the global feature of the assigned class according to the segmentation $\mathbf{K}$.
Second, we replace the \gls{jsd} \gls{mi} estimator with a \gls{dv} one.
Finally, in the last ablation study, we omit our Mutual Information Maximization step and solely perform our Segmentation Step.
As a replacement for our Mutual Information Maximization step we apply the \gls{mi} maximization objective of \gls{iic}, referred to as IIC-MI.
To create the two image versions required by IIC-MI, we use the same transformations as in \gls{iic}.

\Cref{tb:ablation} shows the results of our ablation studies, whereby we performed all experiments using the COCO-Stuff-3 dataset.
We can observe the following:
Using soft assignments increases performance over hard assignments.
A \gls{jsd}-based \gls{mi} estimator performs better than a \gls{dv}-based, which aligns with the results of Hjelm \etal \cite{deepinfo}.
And replacing our Mutual Information Maximization step with the objective of \gls{iic} leads to a decline in performance.

\section{Conclusion}
We proposed a novel approach for unsupervised semantic image segmentation.
%We have shown how we can extend % using local and global features.
Our experiments showed that our method yields semantically meaningful predictions and significantly outperforms related methods.
We used the established datasets for evaluation and introduced a novel challenging benchmark COCO-Person.
Furthermore, we discussed several problems making the quantitative evaluation of unsupervised semantic segmentation challenging.
Finally, we performed ablation studies on our model.
%In particular, too specific, overlapping and underrepresented classes.
%, COCO-persons, As a novel benchmark for unsupervised semantic segmentation we proposed
%a dataset that does not suffer from these problems, COCO-Persons.

\vspace{1.0em}
\small
\noindent
\textbf{Acknowledgements.}
This work was partly funded by the Austrian Research Promotion Agency (FFG) under project 874065.

\bibliographystyle{splncs04}
\bibliography{egbib}

\end{document}